\documentclass[a4paper]{article}
\usepackage{cite}
\usepackage{amsmath,amstext,amsgen,amsbsy,amsopn,amsfonts,amssymb, amsthm}
\usepackage{textcomp}
\usepackage{xcolor}
\usepackage{hyperref}
\usepackage{easybmat}
\usepackage{graphics}
\usepackage[pdftex]{graphicx}
\usepackage{epsfig}
\usepackage{bm}
\usepackage{algorithm}
\usepackage{algorithmic}
\usepackage{balance}

\usepackage{stackengine,scalerel}
\newcommand\dddag{%
  \sbox0{\ddag}\scalerel*{%
  \stackengine{-.6\ht0}{\ddag}{\ddag}{O}{c}{F}{F}{S}}{\ddag}%
}

\begin{document}

\title{A Computational Approach to Improving Fairness in K-means Clustering\\}

\author{
Guancheng Zhou$^{\dag\P}$, Haiping Xu$^{\ddag\P}$, Hongkang Xu$^{\S\P}$, %%\\Yukui Luo$^{\ddag\P}$, 
Chenyu Li$^{\dddag}$, Donghui Yan$^{\dag\P*}$
\vspace{0.12in}\\
$^\dag$Mathematics and Data Science\vspace{0.05in}\\
$^\ddag$Department of Computer and Information Science\vspace{0.05in}\\
$^\S$Charlton College of Business\vspace{0.05in}\\
$^{\dddag}$Data Science, Columbia University\vspace{0.05in}\\
$\P$University of Massachusetts Dartmouth, MA\vspace{0.05in}\\
}

\maketitle

\begin{abstract}
The popular K-means clustering algorithm potentially suffers from a major weakness for further analysis or 
interpretation. Some cluster may have disproportionately more (or fewer) points from one of the subpopulations 
in terms of some sensitive variable, e.g., gender or race. Such a fairness issue may cause bias and unexpected 
social consequences. This work attempts to improve the fairness of K-means clustering with a two-stage optimization 
formulation--clustering first and then adjust cluster membership of a small subset of selected data points. Two 
computationally efficient algorithms are proposed in identifying those data points that are expensive for fairness, 
with one focusing on nearest data points outside of a cluster and the other on highly 'mixed' data points. Experiments 
on benchmark datasets show substantial improvement on fairness with a minimal impact to clustering quality. The 
proposed algorithms can be easily extended to a broad class of clustering algorithms or fairness metrics.
\end{abstract}


\section{Introduction}
\begin{figure}[tb]
\centering
\begin{center}
%%\hspace{-10pt}
\includegraphics[scale=0.36,clip, angle=0]{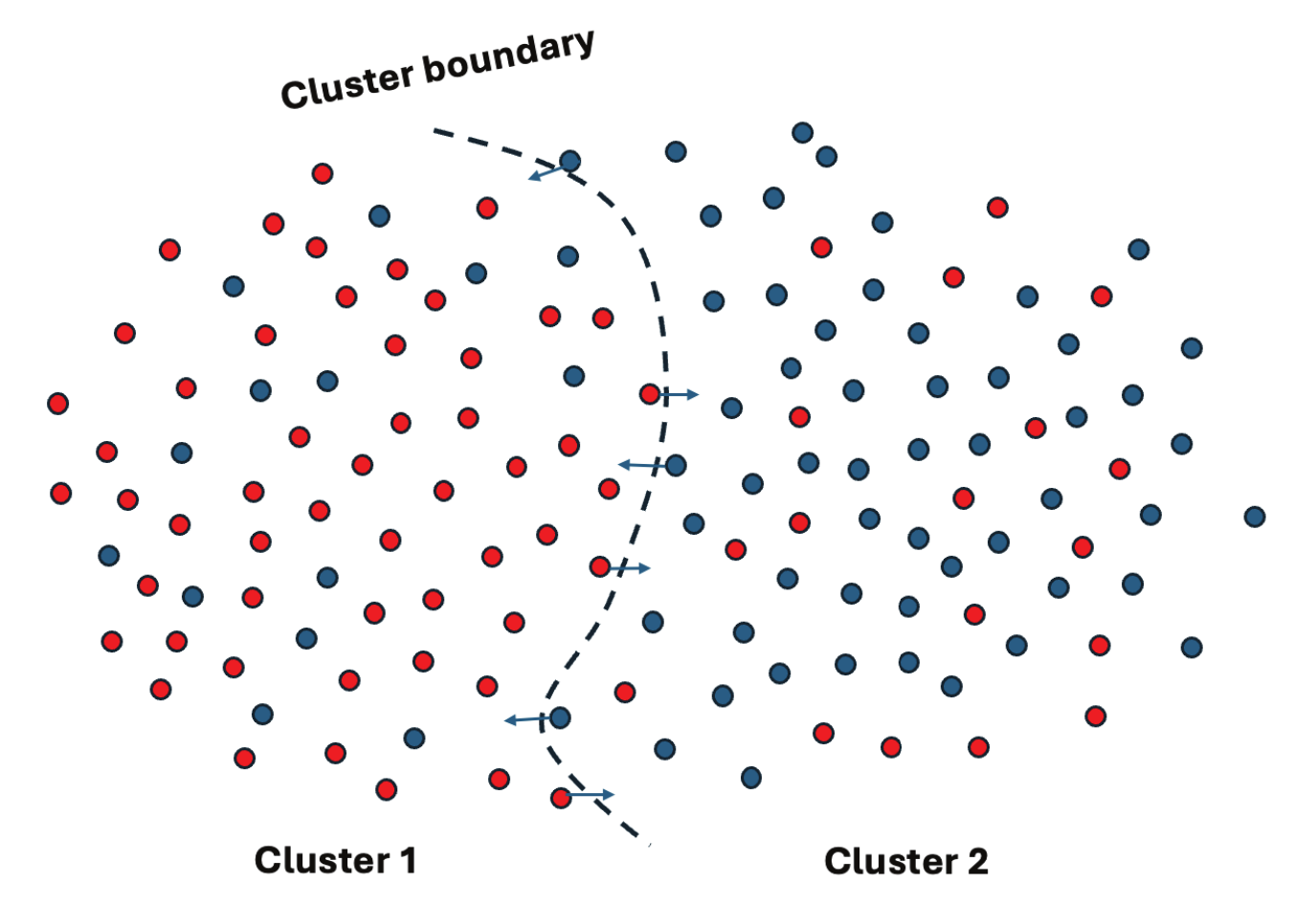}
\end{center}
%%\abovecaptionskip=-1pt
\caption{\it Illustration of the fairness issue in clustering, Points of different color indicate different traits on a sensitive variable,
e.g., gender where blue indicates male and red female. Cluster 1 is dominated by females while Cluster 2 by males. Points with an arrow indicate that we might switch its cluster membership assignment to make the clusters less dominated by one subpopulation.} 
\label{figure:fairnessIllus}
\end{figure}

Clustering is an important problem in data mining. It aims to split the data into groups such that data points in the same group 
are similar while points in different groups are different under a given similarity metric. Clustering has been successfully applied 
in many practical applications, such as data grouping in exploratory data analysis, search results categorization, market segmentation etc. Clustering results are often used for further analysis or interpretation. However, directly applying results obtained from usual clustering algorithms may suffer from fairness issues--some cluster may favor data points from one of the subpopulations, i.e., having disproportionally more points. One example of fairness issue is illustrated in Fig.~\ref{figure:fairnessIllus} where, in each of the clusters, data points coming from one of the subpopulations dominate. Data points in Cluster 1 is dominated by females (marked by red), while Cluster 2 is dominated by males (marked by blue). When the subpopulations of interest are defined over some sensitive features (or variables), such as genders or races, it may cause undesirable bias \cite{SpanakisGolden2013} or social consequences \cite{HeidariFerrari2018, LiuDean2019, PurohitRaghavan2019}. 
\\
\\
This work aims to alleviate the fairness issue arising from clustering, and 
we focus on the popular K-means clustering algorithm \cite{hartiganWong1979} (note that our approach is applicable 
to general clustering with minimal changes to our proposed algorithm). One natural way is to treat clustering as an 
optimization problem, and fairness is viewed as a constraint. However, the constrained optimization problem might be 
computationally hard, even for K-means clustering \cite{ArthurVassilvitskii2006}, due to its nature as a mixed integer 
programming problem. 
While a number of methods have been proposed for fairness in clustering \cite{ChierichettiKumar2017,BeraChakrabarty2019, kleindessnerMorgenstern2019,ChhabraMasalkovaite2021}, they are often very complicated or computationally expensive or tightly
associated with a particular clustering algorithm.
We take a different approach, and work on adjusting the clustering results generated by a given clustering algorithm. We will
identify a small set of `promising' data points and switch their cluster membership assignments to improve the fairness. The 
rationale is that, the inclusion of fairness in the clustering formulation should only change a small part of results in clustering.
%%; of course this set of 
%%points will need to satisfy the condition that switching their cluster membership will not impact too much of the overall clustering
%%quality.  
\\
\\
Our contributions are as follows. We conceive a novel approach to improve the fairness of K-means clustering by adjusting 
the cluster membership of a small selected set of `promising' data points. Two different implementations are proposed, with 
one focusing on nearest data points outside of a cluster and the other on highly `mixed' data points. Both implementations 
are conceptually simple, computationally efficient, and effective in improving the fairness of K-means clustering. Notably,
both are also broad and adaptable in the sense that they are readily applicable to a broad class of clustering 
algorithms or different notions of fairness. 
\\
\\
The remaining of this paper is organized as follows. In Section~\ref{section:method}, we describe our algorithms to improve 
fairness in clustering. This is followed by a discussion of related work in Section~\ref{section:relatedWork}. Experiments and 
results are presented in Section~\ref{section:experiments}. Finally we conclude in Section~\ref{section:conclusion}.
\section{The method}
\label{section:method}
The main idea of our approach is to identify a subset of `promising' data points and switch their cluster membership 
assignments to improve the fairness. Of course, we require that the switching of cluster membership of those selected 
points will not impact much of the overall cluster quality.  
\\
\\
Mathematically, the problem of fair clustering can be formulated as an optimization problem with some cluster quality measure 
as the objective function and an additional constraint on fairness. Let $\mathcal{S}$ be the set of all data points. Let functions 
$g(.)$ and $f(.)$ be the cluster quality measure and fairness measure of interest, respectively. Assuming there are $K$ clusters,
the fair clustering problem is formulated as the following optimization problem $(\star)$
\begin{eqnarray}
\label{equation:kmeansFormu}
&& \underset{\pi} {\operatorname{arg\,min}}~  g(S_1,...,S_K), \label{eq:clusterQ}\\
&& f(S_1,...,S_K) \leq T \label{eq:clusterFair}
\end{eqnarray}
where the clustering operator (i.e., function) $\pi:  \mathcal{S} \mapsto \{1,...,K\}$ maps each data point $x\in  \mathcal{S}$ 
to its cluster index, and $S_i = \{x \in \mathcal{S}:~\pi(x) = i\}$ is the set of data points in the i\textsuperscript{th} cluster, $i=1,...,K$, 
and $T$ is a small tunable parameter. In the above formulation, \eqref{eq:clusterQ} corresponds to the usual clustering 
problem, and \eqref{eq:clusterFair} is a constraint on the fairness defined on clusters obtained by a given clustering $\pi$, i.e., a partition of $\mathcal{S}$ as $ \mathcal{S}= \cup_{i=1}^K S_i$.
\\
\\
Directly solving the constrained optimization problem $(\star)$ can be very costly, as it is a mixed integer programming
problem. Our approach can be viewed as solving $(\star)$ by two stages. In the first stage 
we solve the optimization problem without the fairness constraint \eqref{eq:clusterFair}, i.e., to solve the original clustering 
problem. This leads to a set of feasible solutions (by adjusting cluster membership of promising points) with each corresponding 
to a clustering result (i.e., cluster membership 
assignment) of the best or near-optimal cluster quality measure. The second stage starts from the set of feasible solutions, and 
focuses on the single objective of finding a solution with improved fairness. This is done by switching the cluster membership 
of data points along the cluster boundary. 
\begin{figure}[tb]
\centering
\begin{center}
%%\hspace{-10pt}
\includegraphics[scale=0.36,clip, angle=0]{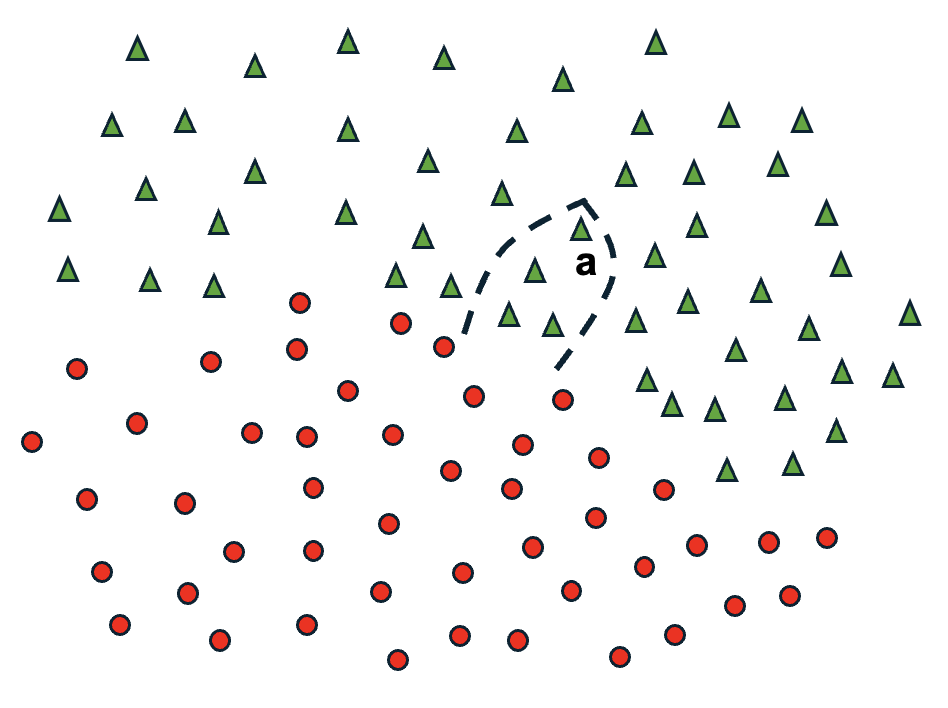}
\end{center}
%%\abovecaptionskip=-1pt
\caption{\it Illustration of connectivity of data points in the same cluster. If one switches point $a$ to the red cluster, then green 
points between it and the red cluster (i.e., points enclosed by the dashed curve) should also be switched.  } 
\label{figure:illusConnectivity}
\end{figure}
\\
\\
Clearly data points that are far away from the cluster boundary should never be touched when re-assigning cluster memberships. 
This is illustrated in Fig.~\ref{figure:illusConnectivity}. Suppose we wish to re-assign the 
cluster membership of one such data point, say $a \in ~\mbox{cluster}~ \mathcal{C}_1$ (i.e., cluster formed by green points) to cluster 
$\mathcal{C}_2$ (i.e., cluster of red points). Then, due 
to the {\it connectivity assumption} (i.e., data points from the same cluster form a ``connected" region), many other data points between 
$a$ and cluster $\mathcal{C}_2$, i.e., data points circled by the dashed circle, should also be switched from cluster $\mathcal{C}_1$ 
to cluster $\mathcal{C}_2$. This will 
lead to a substantial change in the value of the given cluster quality measure and is not desirable. Focusing on the near-boundary data points
allows us to work on a small set of feasible solutions, and computation can be done quickly.
\\
\\
Our definition of fairness takes into account of discrepancy of the proportion of subpopulations in each cluster. For simplicity, assume that there are two subpopulations of interest, and let $p_1$ and $p_2$ be their respective proportions in the entire population. For each cluster $i=1, ..., K$, let $p_{ij}$ be the proportion of data points belonging to subpopulations $j=1,2$. Let $n$ be the total number of data points, and $n_i$ be the number of data points in each of the clusters for $i=1, ..., K$. We can now give the following definition of the fairness index  
\begin{eqnarray}
\label{eq:fairnessIdx}
\mathcal{F} &=& \sum_{i=1}^K  \frac{n_i}{n}   \left( \sum_{j=1}^2 \mid {p_{ij} - p_j}  \mid \right) \nonumber\\
&=& \sum_{i=1}^K  w_i   \left( \sum_{j=1}^2 \mid {p_{ij} - p_j}  \mid \right),
\end{eqnarray}
where term $\left( \sum_{j=1}^2 \mid {p_{ij} - p_j}  \mid \right)$ measures the discrepancy of respective proportions 
of each subpopulation in the i\textsuperscript{th} cluster, and the discrepancies are weighted by the relative size, $w_i$, of each cluster 
out of the entire set of data points. We have $\mathcal{F} \in [0,1]$, and a smaller value of $\mathcal{F}$ would indicate a more fair 
clustering, and $\mathcal{F}=0$ indicates that, in each cluster, all the subpopulations 
have the same proportion as in the general population. Note that our definition can be easily extended to settings 
with multiple subpopulations.
\\
\\
We will start by adjusting points from clusters for which one subpopulation dominates the most. To measure how much 
a subpopulation dominates points from a cluster, say $A$, we define the following cluster balance measure 
\begin{equation*}
\beta (A) = \frac{\left | \{x\in S_1\} \right | }{\left | \{x \in S_2\}\right | },
\label{eq:clusterBalance}
\end{equation*}
where $S_1$ and $S_2$ are the two subpopulations of interest and $| \cdot |$ indicates the cardinality (i.e., number of 
elements) of a finite set. 
Note that the balance measure can also be calculated on the entire dataset (population). If a cluster has a 
large value of cluster balance measure, then that means it has excessive data points from subpopulation $S_1$. On the 
other hand, clusters with a small value of cluster balance index would have insufficient number of data points from 
subpopulation $S_1$. So, two clusters will be better balanced in terms of cluster fairness by exchanging the cluster 
membership of points from a cluster of high cluster balance index with points from clusters of a small cluster balance index.
\\
\\
Having found which two clusters to switch cluster membership for their data points, the remaining issue is 
to identify which data points to adjust cluster membership. The overall idea is to find data points near 
the cluster boundary. Switching the cluster membership of such data points will not impact the cluster quality 
much since they do not contribute much in determining the class boundary as demonstrated by theoretical as 
well as empirical work in classification setting \cite{SinghNowarkZhu2009, transTMA2023}. 
\\
\\
We adopt two different implementations to locate candidate data points for cluster membership swapping, 
while keeping the deterioration to cluster quality `minimal'. One is to pick points that are far away from their 
own cluster centroid but near the centroid of some other cluster, and the other is to pick points that are highly 
mixed in the sense that there are data points from different clusters in a `small' neighborhood of such data points.
These are described in Sections~\ref{section:fcFurtherest}  and \ref{section:fcGini}, 
respectively, after a short description of K-means clustering in Section~\ref{section:kmeans}. 
\subsection{The K-means clustering algorithm}
\label{section:kmeans}
Formally, given $n$ data 
points, $K$-means clustering seeks to find a partition of $K$ sets $S_1, S_2, ..., S_K$ such that the 
within-cluster sum of squares, $SS_W$, is minimized
\begin{equation}
\label{equation:kmeansFormu}
\underset{S_1,S_2,...,S_K} {\operatorname{arg\,min}} \sum_{i=1}^{K} \sum_{\mathbf x \in S_i} \left\| \mathbf x - \boldsymbol\mu_i \right\|^2,
\end{equation}
where $\mu_i$ is the centroid of $S_i, i=1,2,...,K$.
\\
\\
Directly solving the problem formulated as \eqref{equation:kmeansFormu} is hard, as it is an integer 
programming problem. Indeed it is a NP-hard problem \cite{ArthurVassilvitskii2006}. 
The K-means clustering algorithm is often referred to a popular implementation sketched as
Algorithm~\ref{alg:kmeans} below. For more details, one can refer to \cite{hartiganWong1979, lloyd1982}.
\begin{algorithm}
\caption{~~\textbf{$K$-means clustering algorithm}}
\label{alg:kmeans}
\begin{algorithmic}[1]
\STATE Generate an initial set of $K$ centroids $m_1, m_2, ..., m_K$;
\STATE Alternate between the following two steps
\STATE \hspace{\algorithmicindent} Assign each point $x$ to the ``closest" cluster 
	\begin{equation*}     
        \arg \min_{j \in \{1,2,...,K\}} \big \| x-m_j \big \|^2;
        \end{equation*}
\STATE \hspace{\algorithmicindent} Calculate the new cluster centroids
	\begin{equation*}
	m_j^{new} = \frac{1}{\| S_j\|} \sum_{x \in S_j} x, ~~j=1,2,...,K;
        \end{equation*}
\STATE Stop when cluster assignment no longer changes.
\end{algorithmic}
\end{algorithm}
\subsection{Improve fairness by adjusting nearest foreign points}
\label{section:fcFurtherest}
\begin{figure}[tb]
\centering
\begin{center}
%%\hspace{-10pt}
\includegraphics[scale=0.34,clip, angle=0]{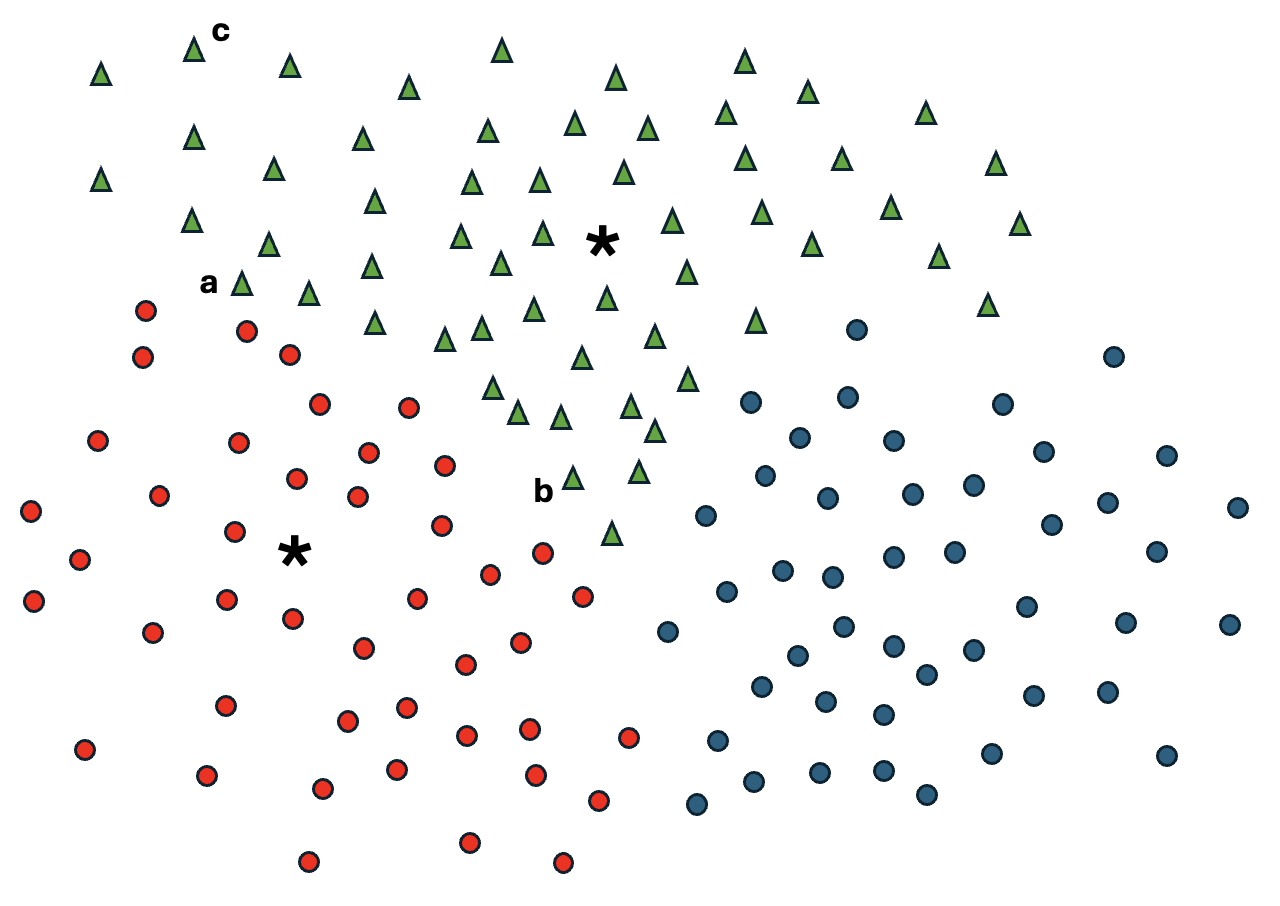}
\end{center}
%%\abovecaptionskip=-1pt
\caption{\it Illustration of the near-foreign heuristic. Points from different classes are indicated by 
different colors. $\star$ indicates the centroid the clusters formed by red points or green points. } 
\label{figure:nearForeign}
\end{figure}

Our first proposed implementation works by switching the cluster membership assignments for data points 
that are far away from their own cluster centroid while near the cluster centroid of other clusters. 
Call this the {\it near-foreign heuristic}. The idea is to switch the cluster membership of data points that 
are less likely to be in their assigned cluster (called source cluster) since they are far away from the cluster centroid
while kind of likely to be part of some other clusters as they are among the nearest data points 
that are outside of the destination cluster (call such points {\it foreign points} of the destination cluster). Since 
such data points are near the cluster boundary, reassigning them will not impact the overall cluster quality much. 
%%
%%
%%\\
%%\\
%%\\
%%
%%
%%
%%
\\
\\
Fig.~\ref{figure:nearForeign} is an illustration of the near-foreign heuristic. Note that simply 
switching the cluster membership of points that are furthest to a cluster centroid will not work, 
as there are two possibilities. One is that the data point is near some other cluster, it makes 
sense to reassign the cluster membership to such points to other cluster which will improve fairness 
while not impacting much to cluster quality. The other possibility is that, although the data points 
are far away from its own cluster centroid, they are not near any other cluster either (rather they 
are near the outmost side of the data space). 
%%\\
%%\\
One example can be seen from Fig.~\ref{figure:nearForeign}, where points $a, b, c$ are all 
far away from the centroid of the cluster formed by green points. If we re-assign points $a, b, c$ 
to the neighboring cluster which is the cluster formed by red points, then clearly point $c$ will not 
make a good choice though points $a$ and $b$ are. Treating data points $a$ and 
$b$ as foreign points of the `red' cluster, then re-assigning them to the `red' cluster is expected, 
and in this case, point $c$ should not be re-assigned to the `red' cluster as it is not among those 
nearest foreign points.  
\\
\\
A description of an algorithm based on the near-foreign heuristic is given as  
Algorithm~\ref{algorithm:fcNearForeign}.  
We say a cluster $A$ is balance enough if its balance measure $\beta(A)$ is close (e.g., within a 
factor of $\beta_0=5 \sim 10\%$) to that calculated on the entire population. Note that the algorithm 
is described for two clusters for simplicity, and we can easily extend it to settings of multiple clusters 
(simply repeat the algorithm for a number of times for a pair of clusters which are highly unbalanced). 
The algorithm has a linear computational complexity. It is upper bounded by the product of the 
number of data points and the number of times we exchange the cluster membership of points for a 
pair of unbalanced clusters.
%%\\
%%\\
%%
%%
%%    
\begin{algorithm}
\caption{~~\textbf{fcNearForeign(X, k)}}
\label{algorithm:fcNearForeign}
\begin{algorithmic}[1]
\STATE Apply K-means clustering algorithm to the data; %%
\STATE Calculate cluster balance measure, $\beta_i, i=1, ..., K$, on each cluster; %%
\STATE Let $A$ and $B$ be clusters with the max and min balance measures, respectively; %% 
\STATE Calculate cluster centroids, $x_A$ and $x_B$, for clusters $A$ and $B$; %% 
\FOR {each data point $X \in A \cup B$}
	%%\STATE Calculate the Gini index for $X$ in its k-nearest neighborhood; 
	\STATE Calculate its distance to cluster centroid $x_A$ or $x_B$ as a foreign point; 
\ENDFOR%%
\STATE Sort such distances in increasing order; 
\STATE Let the respective data points be $X_1, ..., X_m$ with $m=\mid A \cup B \mid$;  
%% 
%%\FOR{i=1 to m}
\WHILE{Clusters $A$ or $B$ not balance enough}
	\STATE Switch the cluster membership for data point $X_i$ w.r.t. cluster A or B; 
	\STATE Clusters $A$ and $B$ now become $A'$ and $B'$, respectively; %%
	\STATE Recalculate the cluster balance measure for clusters $A'$ and $B'$;
	\STATE Stop loop if $A'$ and $B'$ are balance enough; %% 
\ENDWHILE
\STATE Return updated cluster membership assignment; %%
\end{algorithmic}
\end{algorithm} 
\subsection{Improve fairness by Gini index}
\label{section:fcGini}
The Gini index is originally used in tree-based methodology for classification \cite{CART}. 
The classification tree grows by recursively splitting the tree nodes, starting from the root node which consists 
of the entire set of data points, to make the resulting child nodes more ``pure" than their parent nodes. At each 
node split, one looks for a direction and (or) split point such that the overall purity of resulting tree nodes improves 
upon the node split. The Gini index measures how `mixed' or `pure' of points in a node which could be potentially 
from different classes. At a given node, $x$, let the proportion of data points for class $j$ be denoted by $p_j$ for 
$j=1,...,J$, then the Gini index is defined as
\begin{equation}
G(x)=\sum_{j=1}^J p_j (1-p_j).
\end{equation}
%%\\
%%
To appreciate the Gini index as a measure of impurity, let us consider a few numerical examples as illustrated in Table~\ref{table:ToyImpurity}. 
\begin{table}[tb]
\begin{center}
\resizebox{0.48\textwidth}{!}{%
%%\begin{small}
%%\begin{minipage}[b]{0.5\linewidth}%%\centering
\begin{tabular}{c|ccc|c}
\hline
\textbf{Case}                 			& $C_1$     	&  $C_2$     & $C_3$  &Gini  \rule{0pt}{2.4ex}\rule[-0.9ex]{0pt}{0pt} \\
\hline
Case 1  			&1/3			&1/3    &1/3     &2/3 \rule{0pt}{2.2ex}\\
Case 2  			&0.5			&0.30    &0.20  &0.62 \rule{0pt}{2.0ex}\\
Case 3        & 0.80                  & 0.10  &0.10   &0.34 \rule{0pt}{2.0ex}\\
Case 4       & 0.90                  & 0.05  &0.05    &0.185 \rule{0pt}{2.0ex}\\
Case 5  		&1.00   		&0.00 &0.00    &0 \rule{0pt}{2.0ex} \rule[-0.9ex]{0pt}{0pt} \\
\hline
\end{tabular}}
\vskip 10pt
\end{center}
\caption{\it A toy example of node impurity. $C_j$ denote the proportion of points in class $j=1, 2, ..., 5$. Case 1 
and Case 5 are at the two extremes, with Case 1 being the most mixed while Case 5 being the purest. } 
\label{table:ToyImpurity}
\end{table}
Clearly Case 1 is the most mixed while Case 5 is the purest by intuition, and the calculated Gini indices
are consistent with our intuition. 
\\
\\
In this work, we propose a novel use of the Gini index, i.e., use it to detect if a given data point is on or 
near the class (cluster) boundary. The idea is, if a data point is on or near the cluster boundary, then a 
small neighborhood of the data point would likely also consist of data points from different classes (clusters) 
thus a higher Gini index. This will resemble Case 1 or 2. On the other hand, if a data point is in the interior 
of a cluster, then data points from a small neighborhood will mostly be from one cluster, thus a very 
small Gini index. This will be like Case 4--5. Empirically, we can easily find a cutoff value to distinguish 
these two situations. The selection of the right-sized neighborhood is data-dependent, and we make it 
adaptive to the data by using the k-nearest neighborhood \cite{rpForests2019}. The goal is to make the 
neighborhood local and yet large enough to capture the purity of the neighborhood. Empirically we find 
the cluster quality not sensitive to the choice of $k$ (c.f., Fig.~\ref{figure:kappaVarK}).
\begin{figure}[tb]
\centering
\begin{center}
%%\hspace{-10pt}
\includegraphics[scale=0.34,clip, angle=0]{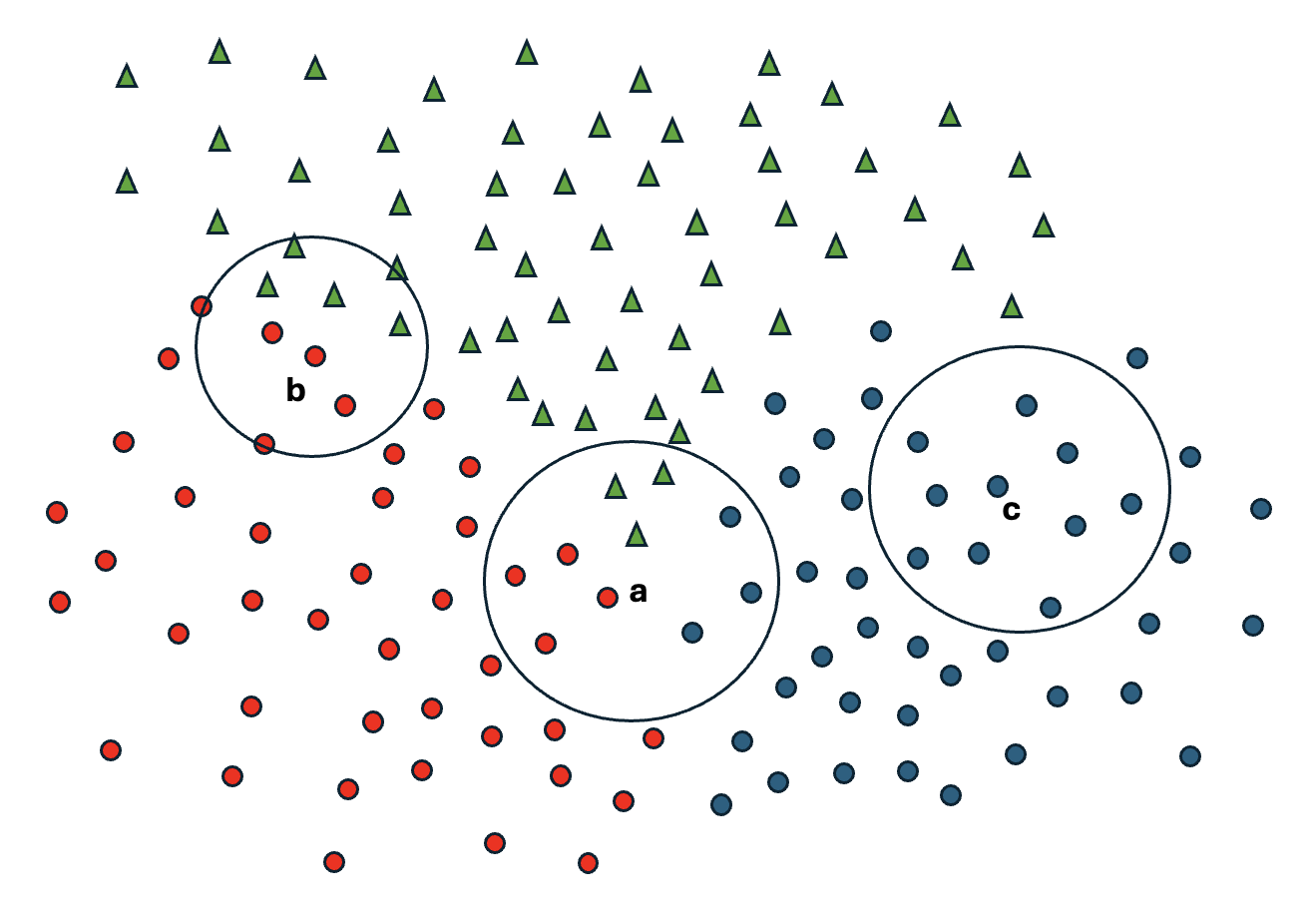}
\end{center}
%%\abovecaptionskip=-1pt
\caption{\it Illustration of boundary and non-boundary points. Points from different classes are indicated 
by different colors. The circle are the 10-nearest neighborhood of given points $a$, $b$ and $c$. The 
respective Gini indices are calculated as $0.66, 0.50$, and $0$ (which indicates the given point is an 
interior point). } 
\label{figure:gini}
\end{figure}
%%
%%
%%
%%\\
%%\subsection{Improving cluster fairness by Gini index}
\\
%%\\
%%
%%
\begin{algorithm}
\caption{~~\textbf{fcGini(X, k)}}
\label{algorithm:fcGini}
\begin{algorithmic}[1]
\STATE Apply K-means clustering algorithm to the data; %%
\STATE Calculate cluster balance measure, $\beta_i, i=1, ..., J$, on each cluster; %%
\STATE Let $A$ and $B$ be clusters with the max and min balance measures, respectively; %% 
\FOR {each data point $X \in A \cup B$}
	\STATE Calculate the Gini index for $X$ in its k-nearest neighborhood; 
\ENDFOR%%
\STATE Sort Gini indices in set $Gini(A \cup B)$ in decreasing order; 
\STATE Let the respective data points be $X_1, ..., X_m$ with $m=\mid A \cup B \mid$;  
\FOR{i=1 to m}
	\STATE Switch the cluster membership for data point $X_i$ w.r.t. cluster A or B; 
	\STATE Clusters $A$ and $B$ now become $A'$ and $B'$, respectively; %%
	\STATE Recalculate the cluster balance measure for clusters $A'$ and $B'$;
	\STATE Stop loop if $A'$ and $B'$ are balance enough; %% 
\ENDFOR
\STATE Return updated cluster membership assignment; %%
\end{algorithmic}
\end{algorithm} 
\\
The proposed algorithm starts by switching the cluster membership assignment for two neighboring 
clusters that are at two extremes in terms of a cluster balance measure. A description of an algorithm 
based on the Gini index is given as  Algorithm~\ref{algorithm:fcGini}. Similarly as that for the Algorithm~\ref{algorithm:fcNearForeign}, it is described for two clusters for simplicity, and we can 
easily extend to multiple clusters. Clearly the algorithm also has a linear computational complexity.    
\section{Related work}
\label{section:relatedWork}
Work related to ours falls into several categories. The first is on algorithms for fair clustering. A number 
of algorithms have been proposed in the last decade, for instance \cite{ChierichettiKumar2017,BeraChakrabarty2019, ChhabraMasalkovaite2021}. Such algorithms generally formulate fair clustering as an optimization problem 
along with an additional constraint on fairness, and then directly solve the optimization problem at its 
entirety. Constraints that involves integer programming would substantially increase the computation. Also, 
such algorithms are typically closely tied to the underlying clustering algorithms. In contrast, our algorithm 
solves the constrained optimization problem in two stages with insights that allow us to focus on a small 
set of feasible solutions, which greatly reduces the problem space at low computational complexity. Also 
our proposed algorithms, though described for K-means clustering, does not explicitly depend on the 
underlying clustering algorithm.
\\
\\
 There are many work that explores various notions of fairness in clustering or other machine learning problems, 
 such as proportional fairness \cite{ChenFain2019, KellerhalsPeters2024}, group fairness \cite{ChengJiang2019}, 
 and individually fair clustering \cite{pmlr-v119-mahabadi20a, pmlr-v151-chakrabarti22a} etc. Also there are work 
 that establishes theoretical guarantees \cite{kleindessnerMorgenstern2019} for fair clustering. Additionally, algorithms 
 have been proposed to account for multiple fairness metrics simultaneously \cite{DickersonZhang2023}.
\section{Experiments}
\label{section:experiments}
\begin{figure}[tb]
\centering
\begin{center}
%%\hspace{-10pt}
\includegraphics[scale=0.4,clip, angle=0]{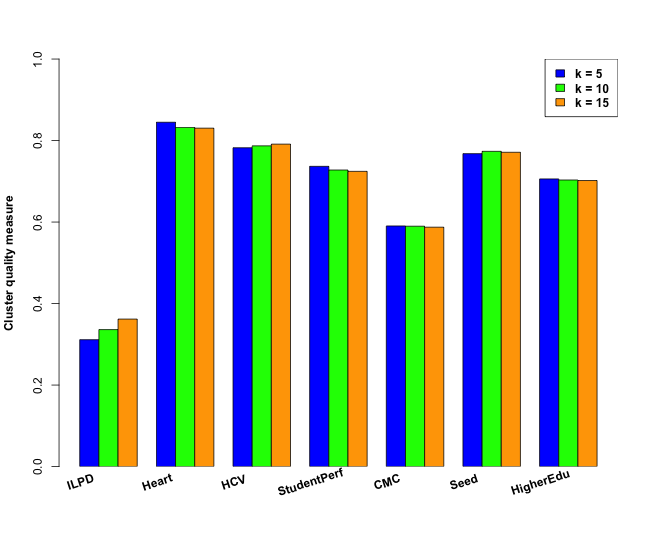}
\end{center}
\abovecaptionskip=-10pt
\caption{\it Insensitivity of the cluster quality over different choices of neighborhood size $k \in \{5, 10, 15\}$ in calculating 
the Gini index of individual data points. The cluster qualities vary very little when $k$ increases from $5$ to $15$. } 
\label{figure:kappaVarK}
\end{figure}

To evaluate the performance of our approach, we choose 7 datasets from the UC Irvine Machine Learning 
Repository \cite{UCI}. This includes the Indian Liver Patient dataset (ILPD), the Heart dataset, the Hepatitis 
C Virus (HCV) for Egyptian patients dataset, the Student Performance (StudentPerf) dataset, the Contraceptive 
Method Choice (CMC) dataset, the Seed dataset, the Higher Education (HigherEdu) Students Performance 
Evaluation dataset. A summary of these datasets is given in Table~\ref{tbl:summaryUCI}.
\begin{table}[tb]
\begin{center}
\resizebox{0.72\textwidth}{!}{%
\begin{tabular}{c|cc|c}
    \hline
\textbf{Data set}     & \textbf{\# Features}  &\textbf{\# instances}  & \textbf{\#Clusters} \rule{0pt}{2.4ex}\rule[-0.9ex]{0pt}{0pt}\\
    \hline
    ILPD  &10     &579              &2 \rule{0pt}{2.2ex}\\
Heart   & 13       &303                &2 \rule{0pt}{2.0ex}\\
HCV    &12		&615                &2 \rule{0pt}{2.0ex}\\
StudentPerf  &30	 &649		     &2 \rule{0pt}{2.0ex}\\
CMC   & 9       &1473                &3 \rule{0pt}{2.0ex}\\
Seed   &7       &210                   &3 \rule{0pt}{2.0ex}\\
%%Computer    &7		&6259                &4\\
HigherEdu  &31	 &145		     &5 \rule{0pt}{2.0ex}\rule[-0.9ex]{0pt}{0pt}\\
\hline
\end{tabular}}
\vskip 10pt
\end{center}
%%\abovecaptionskip=5pt
\caption{\it A short summary of the datasets used in our experiments. } 
\label{tbl:summaryUCI}
\end{table}
\\
\\
For each dataset, we choose one of the variables, such as gender, as the sensitive variable for which 
we will calculate fairness. Note that our proposed algorithms continue to work for fairness defined over 
several variables. We compare the fairness obtained by the original K-means clustering algorithm, the 
{\it fcNearForeign} algorithm and {\it fcGini} algorithm. Two metrics are used for performance 
evaluation in our experiments. One is fairness defined in \eqref{eq:fairnessIdx}, and the other is a commonly 
used cluster quality measure \cite{HTF2001, CF}, defined as
\begin{equation}
\label{eq:kappaDef}
\kappa = \frac{SS_B}{SS_T},
\end{equation}
where $SS_B$ and $SS_T$ are the between-cluster sum of squared distances and the total sum of squared 
distances, respectively. Clearly $\kappa \in [0,1]$ as the calculation of $SS_B$ involves only part of those 
distances used in $SS_T$. A ``large" value of  $\kappa$ would imply that the inter-cluster distances carry 
a large proportion of all pairwise distances, an indication of well-separated clusters thus a good cluster 
quality. Note that here our goal is not to achieve the best $\kappa$, rather to demonstrate that the proposed 
algorithms could improve cluster fairness while not impacting the cluster quality much.   
%%in \eqref{eq:kappaDef}. 
%%
%%
\begin{table}[tb]
\begin{center}
%%\small
\resizebox{0.72\textwidth}{!}{%
\begin{tabular}{c|cc|cc|cc}
\hline
   & \multicolumn{2}{|c|}{$\bm{\mbox{Original}}$}   	& \multicolumn{2}{|c|}{$\bm{\mbox{Near-Foreign}}$}  & \multicolumn{2}{|c}{$\bm{\mbox{Gini}}$} \rule{0pt}{2.4ex}\rule[-0.9ex]{0pt}{0pt}\\
    \hline
ILPD   &0.11 &37.60\%   &0.07     &35.90\%        &0.07   &36.20\% \rule{0pt}{2.2ex}\rule[-0.9ex]{0pt}{0pt}\\
Heart  &0.12 &82.21\%      &0.08 &81.48\%      &0.08 &81.09\% \rule{0pt}{2.0ex} \\
HCV  &0.03 &84.96\%      &0.02 &79.93\%       &0.02 &79.12\% \rule{0pt}{2.0ex} \\
StudentPerf  &0.09  &70.74\%    &0.01 &73.45\%      &0.01 &72.44\% \rule{0pt}{2.0ex} \\
CMC  &0.12  &62.36\%     &0.09 &60.89\%       &0.09 &60.17\%  \rule{0pt}{2.0ex} \\
%%Computer  &0.52  &47.21\%     &0.17 &45.87\%        &0.17  &45.03\%\\
Seed  &0.19    &77.70\%     &0.12   &77.45\%        &0.12     &77.12\%  \rule{0pt}{2.0ex} \\
HigherEdu  &0.32  &75.03\%    &0.27 &71.56\%      &0.27 &70.19\% \rule{0pt}{2.0ex}\rule[-0.9ex]{0pt}{0pt}\\
\hline
\end{tabular}}
\vskip 10pt
\end{center}
\caption{\it Comparison of fairness obtained by the original K-means clustering algorithm, algorithm 
by the near-foreign and the Gini index heuristics, respectively. In each cell, there are two numbers, 
one is the overall fairness index and the other the cluster quality measure $\kappa$.   
}
\label{table:expFC}
\end{table}
\\
\\
Table~\ref{table:expFC} shows that, for all the 7 datasets used in the experiments, our proposed algorithms 
based on both the near-foreign and the Gini index heuristics improve the fairness over the original K-means 
clustering algorithm while maintaining the cluster quality as evidenced in small changes in the value of the 
$\kappa$ measure. It is interesting that, for most datasets, the near-foreign and the Gini index heuristics 
lead to similar cluster quality (i..e, similar $\kappa$ values). It is not clear when one will outperform 
the other. In general, the Gini index heuristic tends to pick those highly mixed data points, that is, points 
that lie at the boundary of several clusters, to switch cluster membership. One weakness with the near-foreign 
heuristic lies in its use of distance in picking data points to re-assign their cluster membership. We have to 
adapt to the right distance metric otherwise performance might be impacted in situations when the geometry 
in the data becomes complex and non-convex, for example, when the data lie on a low-dimensional 
Swiss-roll like manifold in a high dimensional space \cite{RoweisSaul2000,TenenbaumdeSilva2000}.
\\
\\
For the Gini index heuristic, we also assess the sensitivity of cluster quality $\kappa$ w.r.t. the choice of 
neighborhood size $k$ in calculating the Gini index of individual data points. In particular, we vary $k$ over 
$\{5, 10, 15\}$. Fig.~\ref{figure:kappaVarK} shows that the influence to cluster quality is negligible when 
varying the value of $k$. The dataset with the largest impact from $k$ is the ILPD data, and we attribute 
it to the fact that this dataset has mostly integer-(categorical) valued variables. 
\section{Conclusions and future work}
\label{section:conclusion}
We have proposed a novel approach with two implementations to improve the fairness of K-means clustering, one based on 
the near-foreign heuristic and the other based on the Gini index. Both are conceptually simple, 
computationally light and are effective in improving the fairness in K-means clustering algorithms. 
As they do not rely on the implementation of the underlying clustering algorithms, they are readily 
applicable to a broad class of clustering algorithms, for example the popular spectral clustering 
\cite{Ncut, HuangYanNIPS2008, NgJordan2002} and a recently proposed rpf-kernel based clustering \cite{dc22019}. 
Also since our approach solves the fair clustering problem by a two-stage optimization, it can be easily 
adopted to different definitions of fairness. 
\\
\\
While the near-foreign and the Gini index heuristics both explore boundary data points, they are 
different in the sense that the Gini index heuristic attempts to locate those highly mixed data 
points (which may lie at the class boundary of several clusters) which is not necessarily the case 
for that of the near-foreign heuristic. It might be interesting to see what happens if and how one 
may combine these two ideas. Another direction for future work is to extend our algorithms to 
account for multiple fairness measures simultaneously along the line of \cite{DickersonZhang2023}.  
\section*{Acknowledgment}
The work of H. Xu and D. Yan are partially supported by Office of Naval Research (ONR) Grant, MUST IV, 
University of Massachusetts Dartmouth.

%%\section*{References}
%%\begin{thebibliography}{00}

%%\end{thebibliography}

\end{document}